# Large-Scale Text Analysis Using Generative Language Models: A Case Study in Discovering Public Value Expressions in AI Patents

**Sergio Pelaez** *
*School of Public Policy*
*Georgia Institute of Technology*
Atlanta, GA, USA
ORCID: 0000-0003-2640-5623

**Gaurav Verma** *
*School of Computational Science and Engineering, College of Computing*
*Georgia Institute of Technology*
Atlanta, GA, USA
ORCID: 0000-0001-6182-9857

**Barbara Ribeiro**
*SKEMA Business School Université Côte d'Azur, Campus Grand Paris*
Paris, France,
*and*
*Manchester Institute of Innovation Research, University of Manchester*
Manchester, UK
ORCID: 0000-0002-5230-1695

**Philip Shapira** +
*Manchester Institute of Innovation Research, University of Manchester*
Manchester, UK,
*and*
*School of Public Policy*
*Georgia Institute of Technology*
Atlanta, USA
ORCID: 0000-0003-2488-5985

\* Co-First Authors
+ Corresponding author. Email: pshapira@manchester.ac.uk

*Abstract*—Labeling data is essential for training text classifiers but is often difficult to accomplish accurately, especially for complex and abstract concepts. Seeking an improved method, this paper employs a novel approach using a generative language model (GPT-4) to produce labels and rationales for large-scale text analysis. We apply this approach to the task of discovering public value expressions in US AI patents. We collect a database comprising 154,934 patent documents using an advanced Boolean query submitted to InnovationQ+. The results are merged with full patent text from the USPTO, resulting in 5.4 million sentences. We design a framework for identifying and labeling public value expressions in these AI patent sentences. A prompt for GPT-4 is developed which includes definitions, guidelines, examples, and rationales for text classification. We evaluate the quality of the labels and rationales produced by GPT-4 using BLEU scores and topic modeling and find that they are accurate, diverse, and faithful. These rationales also serve as a chain-of-thought for the model, a transparent mechanism for human verification, and support for human annotators to overcome cognitive limitations. We conclude that GPT-4 achieved a high-level of recognition of public value theory from our framework, which it also uses to discover unseen public value expressions. We use the labels produced by GPT-4 to train BERT-based classifiers and predict sentences on the entire database, achieving high F1 scores for the 3-class (0.85) and 2-class classification (0.91) tasks. We discuss the implications of our approach for conducting large-scale text analyses with complex and abstract concepts and suggest that, with careful framework design and interactive human oversight, generative language models can offer significant advantages in quality and in reduced time and costs for producing labels and rationales.

## I. Introduction

This paper introduces a novel approach to text analysis to aid the labeling of complex and abstract concepts. While the method is relevant to different disciplines and studies undertaking text mining and content analysis, we demonstrate its application in the context of science and innovation policy analysis. Specifically, we put forward a semi-automated approach to identifying public values associated with inventions in artificial intelligence (AI) through the analysis of text in AI patent documents. The approach builds on capabilities for classification offered by recently available generative language models (GLMs).

Previous quantitative research in science and innovation policy has focused on the use of bibliometric and text mining methods of sources that include scientific publications and patents. Trends and patterns in research and innovation performance have been discerned using a range of text processing and content analysis techniques (Porter and Cunningham, 2004; Antons et al., 2020). More recently, studies have used BERT-based or similar techniques to categorize research documents according to their respective subject areas based on abstract texts. Examples include tagging publications by their discipline and keywords (Färber and Ao, 2022), identifying AI patent documents (Giczy et al., 2022) and research papers (Sachini et al., 2020), and obtaining multiple category label predictions for social science journal articles (Eykens et al., 2021). A seed and anti-seed set or examples of labels for different categories are required to build a model using these methods. These methods have been applied to large-scale datasets. Qualitative methods have also been used to code and classify unstructured text of scientific publications, patents, and other innovation policy documents (Ribeiro and Shapira, 2020), providing nuanced analyses but with human limitations on the volume of data that can be analyzed.

In this study, we deploy an approach that uses a recently available generative language model to analyze unstructured patent text. In our case, we apply the approach to discover and classify text in AI Patents that conveys attention to public values. It is a semi-automated approach that involves human input and review to identify concepts that are complex and contextually dependent but uses a generative language model and machine learning to accelerate and scale-up classification processes. The approach addresses shortcomings of human-based labeling, such as its high demands on time and resources, yet also supports collaborative annotation and enhances reflexivity in abductive research.

*A. Public Value Expressions in patent documents*

In public policy and administration, public values (PVs) have been characterized as enduring beliefs that are founded on an ideal of human society. As a result, they offer direction, meaning, and legitimacy to collective action (Rutgers, 2015). This concept has been put forward as an improvement over the traditional notion of public interest, which was viewed as ambiguous and lacking practicality as a guide. It was also proposed as an alternative theory to the market failure framework, which was deemed to offer a narrow and sometimes contradictory policy justification. In this sense, PVs are portrayed to be more concrete and practical relative to the idea of public interest while providing broader possibilities for policy deliberation and justification relative to the market failure framework (Bozeman, 2002).

Efforts to identify PVs have involved investigating, for example, scholarly literature or government documents, cultural artifacts, or opinion polls (Bozeman and Sarewitz, 2011). We frame the written articulation of a PV as a public value expression (PVE) where it indicates societal benefits that are promised to or for people, organizations, or ecosystems. A PVE is a signal – it does not mean that the PV will necessarily be realized but it does suggest that there is an idea or intent to do so. Discovering and analyzing such signals, in the context of science and innovation policy, is helpful in understanding potential pathways and directions emphasized and promised by researchers and inventors. At the same time, identifying PVEs presents an operationalization challenge that needs to be addressed if they are to be used as an analytical tool for policy deliberation and justification. PVEs can be difficult to identify as they are contextual, change over time, and require distinctions that are not always evident or easy to discern (Fukumoto and Bozeman, 2019).

Patent documents can be a valuable source of PVEs (Ribeiro and Shapira, 2020). Patents are manifested in written text, where inventors and legal representatives elaborate on their inventions and the context in which they operate. These descriptions provide valuable opportunities to identify and analyze PVEs. Additionally, emerging and general-purpose technologies, such as AI, which are often the subject of patent applications, can serve as a springboard for more comprehensive discussions on society's PVs. Because emerging technologies can change society in fundamental ways it is important to reflect on the extent to which these changes align with our collective values and goals. This can be done by understanding how PVs are mobilized in narratives around these technologies. While doing so, we can gain insights into the intended societal benefits of emerging technologies as well as anticipate their potential negative impacts (Crawford, 2021).

Patent texts are, therefore, valuable materials for studying PVEs because, at one level, they contain detailed descriptions of inventions and the processes used to create them, which can provide insights into the current state of the art, emerging trends, commercial uses, and broader impacts of technologies. At another level, in their background section, patent texts offer clues regarding the social and economic context in which the technology was developed as well as its intended impacts in such a context. The arrival and accessibility of generative language models that can address complex and ambiguous topics, abridge them for human understanding, and organize them effectively, presents an opportunity to explore a new approach to discover PVEs in patent texts. We describe this approach in the following sections, using the case of AI patents.

## II. DATA AND METHODOLOGY

The approach used in this study proceeds through four key stages. First, we create a database of AI documents and identify potential PVEs through keyword filtering. Next, we develop a framework for sentence labeling that involves both human input and AI annotation (using a generative language model). Third, we estimate BLEU scores and perform topic modeling to evaluate the faithfulness, diversity, and discovery capabilities of the AI annotator's output. Finally, we use these annotated labels to train an open-source classifier and apply it to all records. These steps are illustrated in Figure 1 and explained in detail in the following sections.

*A. Data: Obtaining AI patent documents and PVEs*

Our process for obtaining a set of patent documents involves two steps. First, we employ a search strategy developed by Liu et al. (2021) that uses AI-related keywords, cooperative patent classification (CPC) and international patent classification (IPC) codes to retrieve an extensive collection of US patent applications and granted patents filed between 2005 and 2022. To execute the search, we entered a Boolean query into InnovationQ+®, a patent search tool (see https://ip.com/), resulting in the retrieval of patent ID numbers and metadata at the invention level. This yielded a total of 198,456 patent documents, which included a single record for each simple family and granted patents that overrode applications. Second, we used the bulk download option of PatentsView, provided by the United States Patent and Trademark Office (USPTO, see https://patentsview.org/), to extract background, summary, and abstract text. We merged the IDs generated in the first step with the text retrieved in the second step to obtain a final set of 154,934 US patent documents related to AI.

The patent documents were split into individual sentences to facilitate their analysis and categorization through machine learning classifiers. The 154,934 patent documents yielded a total of 5.4 million sentences. To find PVEs within such a large and sparse corpus, we required a method to obtain a manageable subset of sentences with a high density of PVEs to select a sample to annotate. To do so, we implemented a keyword filtering approach. This involved creating a list of single words, bigrams, and trigrams. Initial PV keywords were sampled from a narrative review of relevant documents, both peer-reviewed and non-peer-reviewed, focusing on the benefits and risks of AI. An important part of this literature is reflected in publications related to the governance of this technology. Therefore, we conducted a search for guidelines published by public and private organizations, as well as journal articles analyzing and proposing frameworks for managing the impacts of AI. Based on the initial keywords identified, we looked at possible variations of the same term (e.g., education and educational; equality and equity; sustainability and sustainable) as well as closely related terms (such as well-being and welfare; safety and security; precaution and prevention). Our final list of terms (N=320) represented a broad range of topics related to both the positive and negative impacts of AI, as discussed in the available literature. By filtering sentences containing these terms, we were able to retrieve a smaller subset of sentences (N=73,813) which potentially contained PVEs.

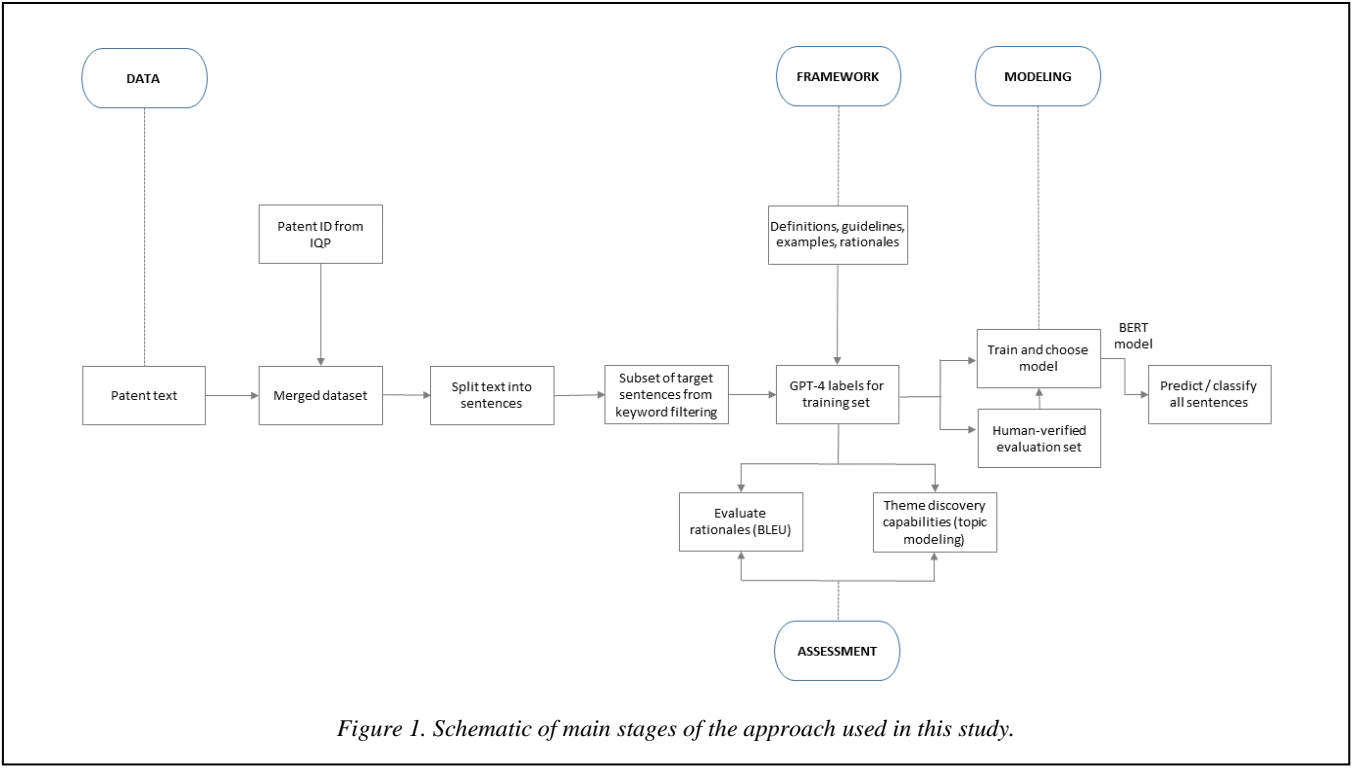

*Figure 1. Schematic of main stages of the approach used in this study.*

From this pool of sentences, reflecting practical and technical factors such as cost, time, and saturation in classification performance, we extracted a training and evaluation set of 10,000 sentences. Instead of randomly drawing sentences, we created a system to rank the importance of PV keywords, grouping them into four categories (1 to 4) based on their ability to capture relevant PVEs. On one end, keywords classified as category 1 were ranked very low in their ability to capture PVEs because they were considered ambiguous. That is, they had the potential to retrieve technical or private value sentences in addition to PVEs. Some examples of category 1 keywords include "risk management" and "emotional state." Additionally, category 1 contained short-tail keywords that retrieved a disproportionately large number of sentences, such as "health care" and "education." Keywords in category 4, on the other end, were characterized by their high ability to capture PVEs. These tended to be long-tail and to retrieve a small number of sentences. Examples of keywords in category 4 are "explainable artificial intelligence," "human safety," "privacy by design," and "societal concern." We used these categories to oversample sentences that were more relevant and undersample those that were less relevant. Specifically, we randomly sampled 4.5% of sentences from category 1, 14% from category 2, 65% from category 3, and 100% from category 4. By doing so, we ensured our training and evaluation sets contained a diverse sample of sentences.

*B. Framework: A method for designing instructions for human and AI annotators*

After obtaining the training and evaluation set, the next step was to generate labels for each sentence, i.e., classify them as PVE (or something else). We developed a multi-stage process to develop a framework for labeling PVEs in patent documents. In the initial stage, we employed an abductive approach – a group of three researchers with experience in text mining of patent documents, PV theory, and emerging technologies went through the sentences and labeled them as either being PVEs or not. We achieved 66%, 56%, and 78% agreement rates in three batches of 50 sentences. The aim was to train ourselves in identifying PVEs in AI patents by observing patterns in disagreements and discussing our labeling decisions. In the second stage, we manually classified sentences on a larger scale, labeling 1,000 sentences with an agreement rate of 79%, leading to a new round of discussion.

A key learning from our discussions was that while the three coders had many areas of agreement about PVEs, in other cases there were differences in the interpretation of PV concepts. This resulted in the rise of differences in labeling sentences. Our individual understanding of PV theories was often implicit and challenging to articulate to one another. Therefore, our labeling decisions were non-standard, and sometimes contradictory. Additionally, in the process of repeated coding, we disagreed with our past selves, failed to be consistent across sentences, and provided varied justifications for our choices. As explained below, we imputed that one of the key reasons for this behavior was the lack of a consolidated paradigm in PV theory, which required us to develop our own framework to support the labeling.

There may be other reasons that generally explain the challenges of labeling complex and abstract topics, such as PVs. In such cases, human labeling capabilities can be diminished by cognitive limitations, such as working memory constraints, mental fatigue from repetitive and lengthy coding tasks, attentional biases, and inflexibility (i.e., once we learn a heuristic to label, we get stuck with it, even if it is wrong). Cognitive biases can also play a role. For instance, our tendency to rely too heavily on initial information (anchoring bias) could result in misclassifying later sentences. The impact of our labeling discrepancies on the final outcome could be particularly significant given that we were only three labelers. Earlier research in fields such as political science, for example, has relied on crowdsourcing to deploy large-scale, qualitative analyses of text to overcome the challenges of bias and inconsistent coding. In this context, the deployment of an increased number of labelers was justified as a way to compensate for individual errors (Benoit et al. 2016).

To overcome the variability of individual coding, we progressed to a third stage, where we provided written justifications for decisions regarding 100 previously labeled sentences. Justifications made our disagreements and the limitations of PV theory more visible. We discussed how PV theory differed from similar frameworks, such as market failure or public interest frameworks. We also debated whether PVE and private value statements were mutually exclusive, the assumptions that were acceptable to infer a PVE from an ambiguous sentence, the role of the target beneficiary in determining whether a sentence was a PVE, the unequal standards to annotate sentences across PV themes, and the consequences of assigning hierarchies to PVEs. Notably, some of these debates are at the center of current discussions in PV theory, as highlighted, for example, in Fukumoto and Bozeman (2019). Our efforts to provide justifications and address contested points led us to create a set of guidelines that encompassed positive heuristics for identifying a sentence as a PVE and negative heuristics for determining when not to label a sentence as a PVE.

These guidelines were initially conceived as a written reference to aid researchers' labeling decisions. The aim was to help us incorporate a deductive approach, where the guidelines were seen as first principles that generalized consistently to observable examples. However, with the introduction in March 2023 of GPT-4 (Generative Pre-Trained Transformer 4, https://openai.com/research/gpt-4), we realized there was an opportunity to use this multimodal large language model to interpret, contextualize, and organize text. We anticipated that our guidelines would be potentially helpful, not only to human annotators but also to AI annotators. The guidelines were both a framework to support human deductive reasoning and a prompt with instructions for a generative language model to understand PV theory and classify patent text accordingly. This led to a fourth stage where, using GPT-4's Application Programming Interface (API), we designed a framework where GPT-4 could label the training and evaluation set. This framework included i) a concrete and comprehensive definition of PV, ii) a categorization of sentences as direct PVE (i.e., the sentence demonstrates that the invention addresses a PV), contextual PVE (i.e., the sentence demonstrates awareness of a PV, but does not provide enough information to link it with the invention), or no PVE, iii) a reduced number of two positive heuristics and three negative heuristics, iv) four examples for each heuristic, v) along with rationales for each example. This framework is included in Appendix 1.

The high-level form of the framework, which most effectively communicates with humans, must be adapted to function as a prompt with instructions for the AI annotator. We provided system instructions to the model to clarify its role and supply necessary definitions to aid categorization. In Figure 2, we show the instructions provided to the GPT-4 model and the role that different components play. Before specifying the definitions crafted as part of the developed framework, we specify the role of the GPT-4 agent (see "Task Specification"). Subsequently, we use the definitions developed as part of our framework (Appendix 1). The definitions are concise, actionable, and can include both inclusionary as well as exclusionary identifiers for each of the categories. Furthermore, based on our early experiments, we added statements to correct the behavior of the generative language model. In the illustrated example, it is to stop the model from making assumptions and inferences based on its inherent knowledge of language (see "Iterative refinement of definitions to correct behavior"). Finally, we note that some definitions can refer to the properties discussed as part of other definitions, as long as the entire set of instructions provided to the model are kept self-contained. As generative language models demonstrate long-form context modeling abilities, they can use the considerations discussed under the definition of "Public Value Expression" to infer "Direct PVE."

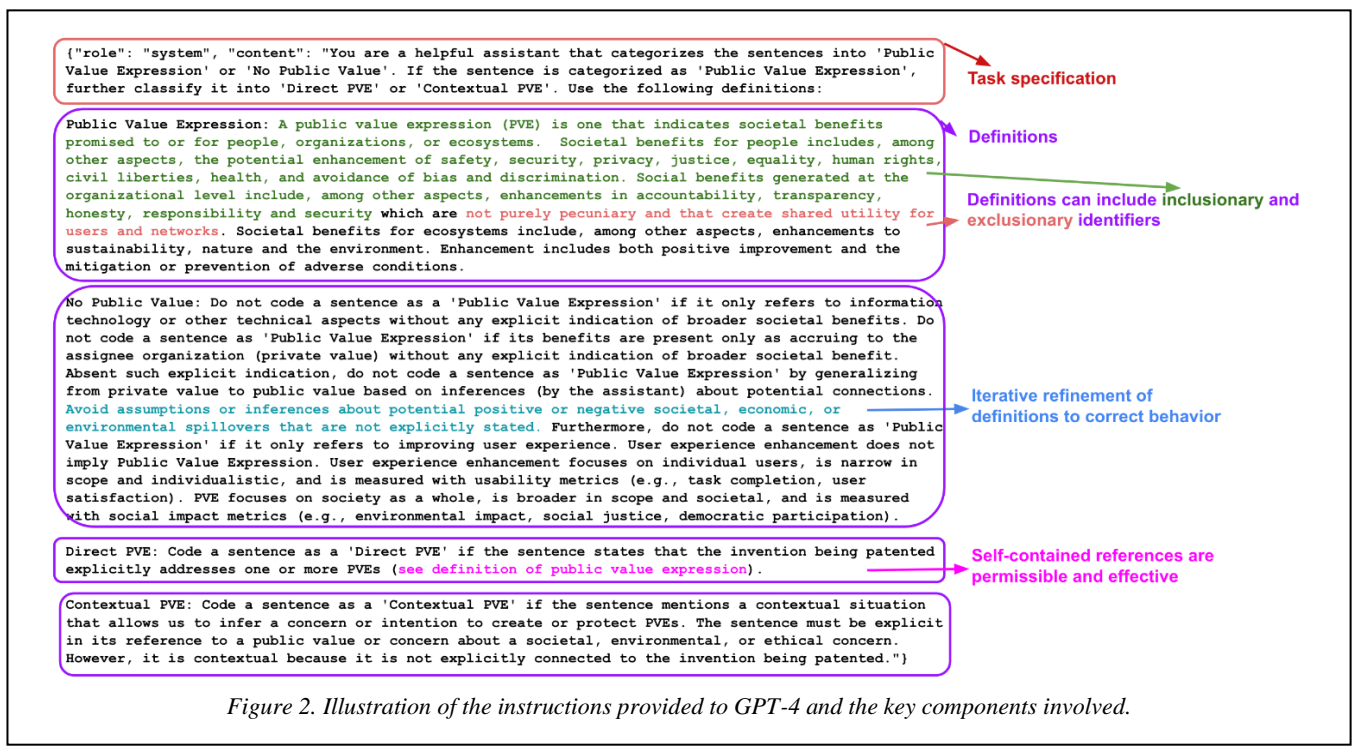

*Figure 2. Illustration of the instructions provided to GPT-4 and the key components involved.*

Following the instructions, we provided 14 examples to GPT-4, spanning the categories under consideration. For each example, we start the rationale with the phrase "Let's think step by step." to trigger chain-of-thought reasoning by GPT-4, a known technique to help the agent arrive at the right prediction following exemplified reasoning scheme. Each rationale follows: "Based on these considerations, I would categorize this sentence as: <final prediction>." The 14 examples were chosen to illustrate a diverse and non-redundant reasoning scheme and were chosen iteratively while analyzing performance on a small set of unseen examples (n = 100). We show one such example in Figure 3.

GPT-4 produced a label (i.e., direct PVE, contextual PVE, or no PVE) and a rationale for its decision. Our agreement with GPT's labels went up to 95% in a separate evaluation set of 1,000 randomly sampled sentences (from the subset of 10,000 sentences). As we demonstrate in the analysis section, not only did GPT produce sensitive labels, but its rationales demonstrated a consistent understanding of our framework and provided logical reasons to support its classifications.

*C. Assessment: Methods for measuring the faithfulness and diversity of the AI annotator's outputs*

In contrast to pre-trained language models such as BERT and its variants, generative language models like GPT-4 can be used for generating coherent long-form rationales along with the final predictions. These rationales serve three purposes. First, chain-of-thought prompting, i.e., encouraging the generative language model to adopt a reasoning scheme by using phrases like "Let's think step by step" before predicting the final label, has demonstrated effectiveness in eliciting an accurate final prediction by the model (Wei et al., 2022). Second, these rationales are material that allows human researchers to assess the labels, making the decision process of the generative language model more transparent and subject to human review. Third, given the high quality of the rationales, these function as persuasive and standardization mechanisms – they help humans in their own labeling decisions and focus their justifications on a smaller set of standard and organized possibilities, thus enabling them to overcome cognitive limitations and biases.

To authenticate this line of argument, we use the rationales that GPT-4 generates while labeling the 10,000 examples to answer the following question: *how diverse and faithful are the rationales (or reasoning mechanisms) and can the generative language model strike a balance between the two measures?*

In prior work, diversity has been measured as the complement of similarity between items (Zhu et al., 2018; Verma et al., 2022; Ma et al., 2010). In this work, we adopt the same approach and measure similarity between two rationales using the BLEU score (Papineni et al., 2002). The BLEU score measures the similarity between two sentences by comparing n-grams (sequences of n words) in a given sentence with the reference sentence. Mathematically,

$$BLEU = BP \cdot e^{\sum_{n=1}^{N} w_n \cdot \log(p_n)}$$

where, $w_n$ are weights for each n-gram, $p_n$ is the precision of n-gram, and N is the maximum order of n-grams used (we compute up to 4 n-grams).

The BLEU score ranges from 0 to 1, where 1 denotes maximum similarity between two sentences. However, since our goal is to measure the diversity of rationales, we interpret the scores as low values demonstrating higher diversity. We are interested in the following forms of diversity: (a) how diverse are the generated rationales with respect to the rationales provided to the model, and (b) how diverse are the generated rationales among each other. For (a), we quantify the category-wise average of maximum similarity that a given generated rationale has with the supplied rationales of the same category. For (b), we take the average of pair-wise similarity scores of all the within-category generated rationales. The same similarity scores can be used to comment on how "faithful" the generated rationales are with respect to the provided rationales. High faithfulness is desired to ensure that the model's reasoning mechanism abides by the same set of reasoning principles as depicted by the supplied rationales.

In addition to using the BLEU similarity scores to assess the quality of the rationales generated by GPT-4, especially regarding striking a balance between diversity and rationales, we assess GPT-4's ability to discover PV themes with topic modeling. To ensure cost-effectiveness, the number of illustrative examples that can be provided to GPT-4 is limited. Therefore, we carefully curate four examples that cover non-redundant themes demonstrating Direct PVEs. However, since these four themes do not cover all possible themes related to PVs, we aim to evaluate whether GPT-4 can produce rationales belonging to different themes not included in the examples. We perform topic modeling on the generated rationales of the subset of sentences labeled "Direct PVE" by GPT-4. More specifically, we use Latent Dirichlet Allocation (LDA) (Blei et al., 2003) to discover 10 topics among all the aforementioned rationales. We chose the number 10 after qualitatively analyzing the coherence and overlap of produced topics. We then qualitatively inspect the 10 topics and assess whether they relate to the themes covered in the provided examples or have been revealed because of the reasoning capabilities of GPT-4.

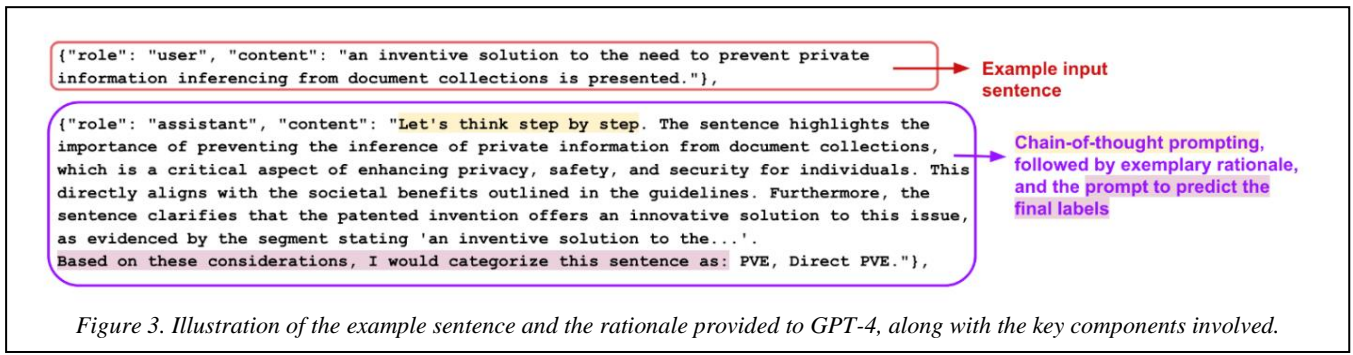

*Figure 3. Illustration of the example sentence and the rationale provided to GPT-4, along with the key components involved.*

*D. Modeling: Training and prediction methods*

The goal of obtaining labels for 10,000 sentences is to train open-source classifiers like BERT (Devlin et al., 2018), which could then be used for inferring the labels of millions of sentences. Given the remarkable performance of GPT-4, which was validated to align with human experts more than 95% of the time, a desirable setting would use GPT-4 to label all the millions of examples in the target corpus. However, using closed-source generative language models for large-scale inference would incur infeasible costs for us (and many other academic research groups). While there are open-source counterparts of proprietary generative language models, like LLaMA (Touvron et al., 2023), OPT (Zhang et al., 2022), and Flan-T5 (Chung et al., 2022), hosting them requires a large GPU infrastructure that may not be accessible to many scholars. Furthermore, current open-source language models are also limited in their maximum context length, which limits the information that can be provided as part of the instructions and examples.

To this end, we deploy a hybrid approach where we use the 10,000 examples labeled using GPT-4 to train open-source pre-trained BERT-like models. BERT, or Bidirectional Encoder Representations from Transformers (Devlin et al., 2018), is a language representation model that uses transformers and bidirectional training to understand the context of words within a sentence. It is pre-trained using a combination of masked language modeling objective and next-sentence prediction on a large corpus of text. Such models can be hosted on a single GPU (NVIDIA A100 80GB, available via platforms like Google Colab) and can be trained for the specific use case at hand with relative ease. For training the models, we use the 9,000 examples labeled using GPT-4 as the training corpus and the 1,000 GPT-identified, human-validated examples as the evaluation corpus. We use two formulations of the classification task: (a) 3-class classification, under which the models are tasked to distinguish between Direct PVE (D-PVE), Contextual PVE (C-PVE), and No-PVE, and (b) 2-class classification, under which the models are tasked to distinguish between PVE and No-PVE. By design, the former setting is more challenging as it involves distinguishing between the sub-categories of PVEs. To quantify the performance of the models, we use macro averages of class-wise F1 scores, precision, recall, and accuracy.

We evaluate a range of pre-trained language models that are similar to the BERT model. We describe these models and the key differences between them. **BERT-base-uncased** is a version of BERT with 110 million parameters. It deals with lower-cased English text, allowing it to be smaller and faster while maintaining robust performance on many tasks. **BERT-large-uncased**, an expanded version of BERT-base, contains 340 million parameters. Despite being computationally heavier, its larger size grants it improved language understanding capabilities. **DistilBERT** (Sanh et al., 2019) is a distilled, lighter version of BERT, maintaining comparable performance while being smaller (66 million parameters), faster, and cheaper to run. It uses a teacher-student training paradigm, learning from a larger "teacher" BERT model. **RoBERTa-large** (354 million parameters), or Robustly Optimized BERT, is a variant of BERT that modifies BERT's training process for improved performance, using larger batches of data, more data, and removing the next-sentence prediction task (Liu et al., 2019). **ALBERT-xxl-v2** (223 million parameters), or A Lite BERT, is a version of BERT that introduces parameter-reduction techniques to lower memory consumption and increase training speed. It achieves this by sharing parameters across layers and factorizing the embedding layer (Lan et al., 2019). Finally, **DeBERTa-xxl-v2** (1.5 billion parameters), or Decoding-enhanced BERT with disentangled attention, improves upon BERT and RoBERTa with a disentangled attention mechanism (He et al., 2020). This allows the model to dynamically integrate contextual information, improving its understanding of complex language phenomena.[1]

To provide context for the performance of the classifiers obtained from training these pre-trained language models, we also include the performance that random classifiers would achieve on the evaluation set. These random classifiers, uniform and biased based on class ratio, serve as a baseline comparison for evaluating the effectiveness of our trained classifiers.

### III. ANALYSIS

*A. Evaluating GPT-4 rationales*

Table 1 shows three examples of sentences from patent texts, their labels, and the sentences generated by GPT-4. These examples illustrate how GPT-4 interprets the patent text and provides rationales for its decisions. The rationales reveal the following steps that GPT-4 takes: (a) it checks for an explicit reference to a PV in the sentence, (b) it identifies possible PV themes that are related to the reference, and (c) it looks for evidence that connects the patented invention and the PV (e.g., "an objective of the present invention…"). In addition, GPT-4 considers the negative heuristics that we instructed it to use, as seen in the rationale of the "No PVE" sentence, where it explains that the sentence focuses on the technical aspects of the invention rather than on a PV. In other "No PVE" sentences, GPT-4 states that enhancing user experience or creating private value does not necessarily imply contributing to PVs.

---

[1] We also experimented with BERT for patents (Lee et al., 2019), which was trained on 100 million patents and has demonstrated notable performance in tasks like searching for prior art while drafting patents, autocomplete, and automatically generating classification codes. Therefore, one might expect BERT for patents to be among the models being used for this task. However, we observed that the performance of BERT for patents is lacking (F1 score of 0.78 for the 3-class classification task) in comparison to the other models mentioned here (F1 score > 0.80 for the 3-class classification task). We believe that this is because identification of public values is not a conventional patent-related task and requires the understanding of broader context, which domain-specific models like BERT for patents may lose because of restrictions to the training corpus.

TABLE 1. EXAMPLES OF GPT-4 RATIONALES FOR PVE LABELING

| Sentence | Humanly verified GPT-4 label | GPT-4 rationale |
|---|---|---|
| It is further an objective of the present invention to assess the risk of a service provider engaging in collusion with one or more partners in the ecosystem of service providers. | Direct PVE | "The sentence explicitly states that the invention aims to assess the risk of collusion among service providers within an ecosystem. This aligns with the societal benefits for organizations, as it enhances transparency, honesty, and responsibility, which are not purely pecuniary and create shared utility for users and networks. The invention is explicitly identified as addressing this issue, as indicated by the segment of the sentence that reads 'it is further an objective of the present invention to…'". |
| In all cases, data security is a priority due to the sensitive nature of medical information. | Contextual PVE | "The sentence explicitly highlights the importance of data security in the context of medical information, which is a critical aspect of enhancing privacy, safety, and security for individuals. This directly aligns with the societal benefits outlined in the guidelines. However, the sentence does not explicitly connect this priority to a specific invention or patent. |
| In marked contrast are artificial neural networks, which have been disparagingly referred to as "black boxes" for their inability to model a problem in an intuitive fashion. | No PVE | The sentence discusses the limitations of artificial neural networks, referring to them as "black boxes" due to their lack of intuitive problem modeling. However, it does not explicitly or implicitly mention any broader societal benefits or public values. The focus is on the technical aspects of artificial neural networks rather than their potential impact on society." |

To evaluate the quality of GPT-4's rationales, we can use an additional method based on BLEU scores. Our goal is to ensure that these rationales are both faithful and diverse. This means that our framework should serve as a reliable guide for generating rationales for unseen examples. Additionally, the generative language model should be able to deduce the label of any PV theme – even if it is not included in the examples -- by leveraging the first principles outlined in the framework.

We analyze the diversity of the GPT-4 rationales with respect to the illustrative examples supplied to GPT-4. Recall that a lower BLEU score indicates more diversity. We use BLEU scores (see Table 2) to quantify the pairwise similarity among within-category rationales provided to GPT-4 (first column after categories), the average maximum similarity of generated rationales with within-category provided rationales (second column), and the average pairwise similarity among within-category generated rationales (third column).

TABLE 2. DIVERSITY AND FAITHFULNESS OF RATIONALES

| Category | Pairwise similarity among within-category rationales | Average maximum similarity with provided rationales | Pairwise similarity among generated rationales |
|---|---|---|---|
| NO-PVE | 0.02 | 0.25 | 0.23 |
| Direct-PVE | 0.04 | 0.16 | 0.07 |
| Contextual-PVE | 0.08 | 0.20 | 0.13 |

As indicated, we first looked at the diversity of the rationales that were provided, and the average of pairwise similarity among within-category rationales was calculated. These low values indicate that the provided rationales are very diverse with respect to each other, which is expected as the examples were chosen to be non-redundant. Next, we analyzed the diversity of the generated rationales with respect to the provided rationales. The average maximum similarity between each generated rationale and the provided rationales for the same label was calculated. The increase in BLEU scores indicates that the provided rationales are being used as guidelines for generating the rationales of unseen examples, a behavior often referred to as faithfulness. However, it is worth noting that the BLEU scores are not exceptionally high, which demonstrates diversity in the sampled data and the rationales required to categorize the examples. Finally, we looked at the diversity of the generated rationales with respect to each other. The average of pairwise similarity among within-category rationales was calculated. The observed results are expected, as the generated rationales are anchored in the provided rationales and show some similarity but not too much. The diversity among Direct-PVE is notably higher in comparison to Contextual-PVE and NO-PVE. One possible explanation for why Direct PVEs are more diverse than Contextual or NO PVEs is that Direct PVEs are more likely to reflect the specificity and novelty of the AI invention and how it addresses a societal challenge or benefit. In contrast, contextual or no public value expressions are more likely to reflect the generality and commonality of AI technologies and their potential applications.

We also assess GPT-4's ability to discover PV themes through topic modeling. We provided 4 examples to GPT-4 for the category "Direct PVE," which spanned the following themes: security and privacy, disease risk prediction, occupational health, and exiting poverty. To assess whether GPT-4 rationales cover themes beyond the ones provided in the examples, we use topic modeling to identify 10 themes in the rationales produced for 1,258 unseen sentences labeled as "Direct PVE." With qualitative analysis of frequently occurring words within each topic, we found the following 10 topics. We also indicate whether these topics are "covered" in the provided rationales or are "revealed." We note that we adopt a very conservative approach while considering a rationale as "revealed" only considering topics completely unrelated to the topics provided in examples.

- Topic 1: Security and safety enhancement (**covered**)
- Topic 2: Mental health, well-being, patient, and medical monitoring (**covered**)
- Topic 3: Privacy, data security, and personal information protection (**covered**)
- Topic 4: Clean energy, sustainability, and renewable resources (**revealed**)
- Topic 5: Economic and financial development (**revealed**)
- Topic 6: Transparency in decision-making, machine bias, and participation consent (**revealed**)

- Topic 7: Developing organizational trust among people (**revealed**)
- Topics 8 and 9: Modeling and assessment of disease risk (**covered**)
- Topic 10: Applications for assessing human behavior (**revealed**)

We note that GPT-4 rationales for categorizing content as Direct PVE cover some of the new themes that are unrelated to themes covered in the examples (4 out of 10), whereas a majority of them are related to the ones that have been included in the examples (6 out of 10). These results further advance the argument that GPT-4 can strike a balance between faithfulness with the provided rationales while exhibiting reasoning based on "revealed" themes.

Based on the above two tests, using BLEU scores and topic modeling, the results indicate that (a) GPT-4 has balanced diversity and faithfulness in justifying its PV classification decisions, and (b) GPT-4 internalized our framework and used it as a deductive device to classify the various types of PVEs it encountered.

### B. Training open-source classifiers

In the final major step of our approach, we use the labels produced by GPT-4 to train open-source classifiers. We report the results for the two settings – three-way classification and two-way classification in Tables 3 and 4. First, we note that the performance for both the classification settings is satisfactory – the best models achieve an F1 score of almost 0.85 for the 3-class classification task and an F1 score of above 0.90 for the 2-class classification task. This means that the models were able to learn with the labels obtained from GPT-4 and predict accurately on a held-out evaluation set. Second, we note that the fine-grained classification task is relatively more difficult for the BERT-based classifiers, as it involves distinguishing between Contextual and Direct PVEs. Finally, we observe a clear trend across the two tasks: the number of parameters in the pre-trained language model positively affects the classification performance. We can note that larger models (DeBERTa-xxl-v2 with 1.5 billion parameters) perform better than models with fewer parameters. This pattern aligns with the scaling law suggesting that substantial performance improvements are achieved by scaling up LLMs (Bowman, 2023).

### DISCUSSION AND CONCLUSIONS

The availability of labeled data is an important requirement for training language classifiers for subsequent large-scale machine-based text analysis. However, the manual labeling of text is a resource-intensive and time-consuming process, limiting the volume of data that can reasonably be coded by humans. Additionally, when dealing with complex and abstract concepts such as public values, manual labeling also carries cognitive limitations and biases that can reduce the quality of the labels. To address this problem, this paper puts forward an alternative semi-automated approach where a generative language model serves as the main annotator, with human validators providing verification on smaller samples. This approach requires crafting a comprehensive and intelligible framework with guidelines on what to do (i.e., positive heuristics), what not to do (i.e., negative heuristics), precise definitions, examples, and rationales. Such a framework acts as a prompt with instructions for the AI annotator.

TABLE 3. CLASSIFICATION PERFORMANCE OF MODELS TRAINED ON DATA LABELED USING GPT-4, 3-CLASS CLASSIFICATION (NO-PVE, D-PVE, AND C-PVE)

| Model | # Parameters | F1 (Macro) | Precision (Macro) | Recall (Macro) | Accuracy |
|---|---|---|---|---|---|
| Random (uniform) | – | 0.3159 | 0.3357 | 0.3454 | 0.3370 |
| Random (biased) | – | 0.3336 | 0.3337 | 0.3336 | 0.4260 |
| DistilBERT-base-uncased | 66 millions | 0.8107 | 0.8124 | 0.8091 | 0.8370 |
| BERT-base-uncased | 110 millions | 0.8210 | 0.8207 | 0.8221 | 0.8480 |
| BERT-large-uncased | 340 millions | 0.8149 | 0.8236 | 0.8072 | 0.8510 |
| RoBERTa-large | 354 millions | 0.8373 | 0.8464 | 0.8291 | 0.8630 |
| ALBERT-xxl-v2 | 223 millions | 0.8343 | 0.8435 | 0.8269 | 0.8570 |
| DeBERTa-xxl-v2 | 1.5 billions | 0.8414 | 0.8518 | 0.8351 | 0.8617 |

TABLE 4. CLASSIFICATION PERFORMANCE OF MODELS TRAINED ON DATA LABELED USING GPT-4, 2-CLASS CLASSIFICATION (NO-PVE AND PVE)

| Model | # Parameters | F1 (Macro) | Precision (Macro) | Recall (Macro) | Accuracy |
|---|---|---|---|---|---|
| Random (uniform) | – | 0.4983 | 0.5001 | 0.5001 | 0.4990 |
| Random (biased) | – | 0.5397 | 0.5397 | 0.5398 | 0.5430 |
| DistilBERT-base-uncased | 66 millions | 0.8107 | 0.8124 | 0.8091 | 0.8370 |
| BERT-base-uncased | 110 millions | 0.8716 | 0.8861 | 0.8701 | 0.9042 |
| BERT-large-uncased | 340 millions | 0.8901 | 0.8993 | 0.8867 | 0.9114 |
| RoBERTa-large | 354 millions | 0.9013 | 0.9136 | 0.9004 | 0.9218 |
| ALBERT-xxl-v2 | 223 millions | 0.9046 | 0.9172 | 0.9081 | 0.9278 |
| DeBERTa-xxl-v2 | 1.5 billions | 0.9103 | 0.9256 | 0.9129 | 0.9310 |

The quality of the framework's output depends on both the written content and the capabilities of the generative language model. We refined our definitions, guidelines, and examples through multiple iterations to reduce the error rate of GPT-4's results. We note that while our framework design worked well with GPT-4, it had limitations with GPT-3.5. When we tested the same framework on both models, GPT-4 produced outstanding results, while GPT-3.5 showed only a slight improvement over human labeling. Other text-based experiments have found a similar contrast between these two versions of GPT (Nori et al., 2023). GPT-4 appears to be already capable of addressing complex classification tasks, and future versions will likely require less framework design work and provide more accurate labels and sophisticated rationales for abstract concepts (Bubeck et al., 2023).

In our framework, we used GPT-4 to label a training set (at cost) which was then used to predict within a much larger set using open-source classifiers. We found that the labels generated using this approach were consistent and accurate, with convincing rationales that aided humans in organizing their ideas and differences around intricate topics. The generative language model produced rationales that followed the instructed guidelines (i.e., it is faithful) while generalizing to label unseen topics (i.e., it is diverse). Our analysis of the rationales offers insights into the reasoning capabilities of generative language models, demonstrating that they can strike a balance between faithfulness and diversity when prompted with carefully crafted instructions and examples.

To be clear, the approach presented in this paper was not costless. The charge for labeling the 10,000 sentences in March 2023 via the Open-AI GPT-4 API was about US$ 1,200, with just under 40 million tokens used. However, the cost in terms of time and money was far less than the fully manual coding of 10,000 sentences. Generative language models may become less expensive to use or increasingly available for large-scale language processing on an open-access basis. With the support of generative language models, the semi-automated framework that we put forward in this study is likely to be useful in many use cases, especially when analyzing complex concepts. In particular, the approach has the potential to be widely useful in the social sciences, including in science and innovation policy analysis.

A key element of our approach is human-machine interaction. The capabilities of the human side are important. We recognize that the rise of generative language models raises concerns about the risk of bias, even in classifying text. Complex and abstract concepts like PVs require training in and understanding of public administration, policy, responsible innovation and ethics, and related fields, meaning that, unlike labeling examples where it is easy to discriminate between categories (e.g., an apple is a fruit and an iris is a plant), complex and abstract concepts cannot be left for outsourced labeling (e.g., Mechanical Turk). On the machine side, even with more capable generative language models, it would be a mistake to assume that human labor is no longer necessary for large-scale text analysis. On the contrary, it is more critical than ever. Human researchers are indispensable in designing frameworks, engineering prompts, and validating results. Crucially, the careful development of training instructions depends on a series of human-led steps to ensure the quality of the analytical lens being constructed. Further, human supervision of intermediary outputs is equally necessary. In their supervised machine learning approach to management analysis, Harrison et al. (2022), show how different phases of the process, from construct identification to data scaling, depend on human manual labor. This is in line with the findings of recent studies looking at the impact of digitalization on scientific work (Ribeiro et al. 2023). As this research demonstrates, these activities remain a time- and intellectually-intensive task that requires the expertise and creativity of skilled professionals.

Despite the fact that we deploy a framework for AI-assisted analysis of text, the depth of such part of the analysis remains largely descriptive, in content analysis terms (Krippendorff 2019). It is through human iterations with the text, with each other and, crucially, with AI through its labeling and justifications behind its choices, that the analysis becomes more interpretive and subjective. This movement between human and AI labeling resembles hybrid approaches adopted elsewhere through collaboration between computational and qualitative researchers (Liu et al. 2021). We argue that using generative language and machine learning models can both enhance and restrict the subjectivity component of content analyses and qualitative research. Ultimately, the descriptive or interpretive tendencies of the analysis will depend on the context of each study, including its research questions, disciplinary orientations, the characteristics of the theoretical frameworks with which researchers are engaging, and the shortcomings of AI models in dealing with more subjective interpretations of the data.

The approach used in this study builds on the strengths of generative language models in classification and pattern identification. There are limitations which should be kept in mind when interpreting the results. Our empirical data is restricted to text found in US AI patent documents. The expression of PVs in that text depends on the motivations of the inventors, assignees, and agents involved in preparing these documents. Our interpretation of PVs is collated from a comprehensive review of PV discussions in available English-language peer-reviewed publications and other documents. Broader or narrower interpretations of what is a PV could be posited. The semi-automated approach to using general language models has a cost and requires coding skills, as well as research methodology skills.

In work already under way, and aware of these limitations, we will use our approach to labeling and organizing to analyze trends and patterns of PVE expressions in AI patent documents. We will focus attention on the drivers of the inclusion of PVEs and the potential societal and policy implications of the presence (and absence) of PVEs in emerging AI technologies. We anticipate that approaches similar to those described in this study to use generative language models for labeling and organizing complex and abstract concepts, as well as classifying text at large scales, will be taken up by other researchers. While our specific study found substantial improvements over manual coding, there is a need for further research to compare and validate the use of generative language models in classifying complex social science concepts in texts. However, our initial experience with embedding a semi-automated generative language model approach is promising; the findings suggest significant potential to assist researchers – in science and innovation policy analysis as well as other domains – in text labeling and organization.

ACKNOWLEDGMENTS

This work was supported in part by the Partnership for the Organization of Innovation and New Technologies, SSHRC [grant number 895-2018-1006] (SP, PS); and the Biotechnology and Biological Sciences Research Council [grant number BB/W013770/1] (PS). GV is partially supported by the Snap Research Fellowship.

## APPENDIX 1

These Guidelines for Public Value Expressions (developed by the authors) are used to develop prompts for the generative language model.

(Text in italics is drawn from selected AI patents in the study dataset; bold text is emphasis added by the study authors.)

*A. Public Value Expression*

A public value expression (PVE) is one that indicates societal benefits promised to or for people, organizations, or ecosystems.[2] Societal benefits for people includes, among other aspects, potential enhancement of safety, security, privacy, justice, equality, human rights, civil liberties, health, and avoidance of bias and discrimination. Social benefits generated at the organizational level include, among other aspects, enhancements in accountability, transparency, honesty, responsibility and security which are not purely pecuniary and that create shared utility for users and networks. Societal benefits for ecosystems include, among other aspects, enhancements to sustainability, nature and the environment. Enhancement includes both positive improvement and the mitigation or prevention of adverse conditions.

*B. Guidelines for coding sentences as PVEs in AI Patents*

**P. Code as a PVE**

**P1**. Code a sentence as a PVE if the sentence states that the invention being patented explicitly addresses one or more PVEs (see definition of public value expression). (Direct PVE).

*P1 examples.*

- *"an **inventive solution** to the need **to prevent private information inferencing** from document collections is presented."*
  - o Rationale: The sentence highlights the importance of preventing the inference of private information from document collections, which is a critical aspect of enhancing privacy, safety, and security for individuals. This directly aligns with the societal benefits outlined in the guidelines. Furthermore, the sentence clarifies that the patented invention offers an innovative solution to this issue, as evidenced by the segment stating, "an inventive solution to the...".

- *"a **disease risk prediction method** according to the **present invention** includes: predicting a development risk of an infectious disease using a prediction model for predicting a development status of the infectious disease, the prediction model being learned based on electronic data of a patient; and outputting the predicted development risk."*
  - o Rationale: The sentence clearly states that the invention being patented addresses the need for disease risk prediction, which falls under the potential enhancement of health for individuals. The segment of the sentence that says "a disease risk prediction method according to the present invention" identifies that the patented invention is the one addressing this issue.

- *"yet another **object of the disclosure** is to provide a system and method to characterize **risk of occupational health hazards** and implement control **measures to protect workers**, with real time evaluations of environmental conditions and exposure concentrations that provides a guide through a task exposure risk assessment."*
  - o Rationale: The sentence clearly states that the invention being disclosed aims to improve occupational health by characterizing risks and implementing control measures to protect workers. This aligns with the guidelines' definition of potential enhancements in health, as well as accountability or security that benefit users and networks beyond monetary gain. Therefore, the invention promises a societal benefit for individuals and organizations. We can identify the patented invention as the one addressing this issue, as indicated by the segment of the sentence that reads "yet another object of the disclosure is to provide a system and method to…".

- *"brief **description of the disclosure** in one aspect, **a computer-implemented method** for selecting participants from an applicant pool to participate in a program to **exit poverty** is described."*
  - o Rationale: The sentence describes a computer-based method for selecting participants from an applicant pool to exit poverty, which is a societal benefit for individuals and organizations. The invention is explicitly identified as addressing this issue, as the sentence begins with a brief description of the disclosure and then explains how the invention benefits society.

**P2.** Code a sentence as a PVE if the sentence mentions a contextual situation that allows us to infer a concern or intention to create or protect PVEs. The sentence must be explicit in its reference to a public value or concern about a societal, environmental, or ethical concern. However, it is contextual because it is not explicitly connected to the invention being patented (Contextual PVE).

*P2 examples.*

- *"**Social robots** may be used to augment **teaching** and may have particular efficacy in working with **autistic children** as well as in providing **companionship to elderly persons** who live alone."*
  - o Rationale: The sentence is explicit in its reference to a public value, which is the potential benefit to both autistic children and elderly people who live alone. The sentence suggests that the use of social robots can enhance teaching and provide companionship to these two vulnerable groups. Therefore, the sentence indicates a concern for the enhancement of the quality of life and well-being of individuals who are otherwise disadvantaged or isolated. However, the connection between the patented invention and public value is not explicit in the sentence. The sentence is only contextual.

---

[2] This adapts the three categories of potential AI harms (to people, organizations, and ecosystems) identified in the NIST AI Risk Management Framework (January 2023). https://www.nist.gov/itl/ai-risk-management-framework.

- *"such an **enabled machine** would be of an immense assistance to the **development of human civilization** much further and much faster leading to **abundance, economic prosperity, biological and mental health, and well-being of society**."*
    - o Rationale: The sentence is explicit in its reference to a public value, which is the potential benefit to society as a whole, including abundance, economic prosperity, biological and mental health, and well-being. The sentence suggests that an enabled machine could significantly contribute to the development of human civilization, leading to broad societal benefits. However, the connection between the patented invention and public value is not explicit in the sentence. The sentence is only contextual.

- *"The natural ecology of many **aquatic environments** around the world is under **threat** due to **human activity, climate change**, and other sources."*
    - o Rationale: The sentence highlights the potential danger that human activity, climate change, and other sources pose to the natural ecology of aquatic environments globally. While it does not explicitly mention any actions or intentions to address this threat, the sentence conveys a sense of concern for the well-being of ecosystems.

- *"**malicious network information has seriously impacted** and threatened individuals, enterprises and even countries, as well as **safety and social public interests**."*
    - o Rationale: The impact of malicious network information on individuals, enterprises, and even countries highlights the need to protect safety and social public interests. Although the sentence does not specify any particular actions or intentions to address this concern, it implies a recognition of the importance of safeguarding the safety and security of individuals and society.

**N. Not a public value expression**

**N1**. Do not code a sentence as a PVE if it only refers to information technology or other technical aspects without any explicit indication of broader societal benefits.

*N1 examples.*

- *in some embodiments, the first sensor collects a real-time actual environmental condition at the first location of interest as the first source of sensor data, and the second sensor collects information that is a possible impact on the first source of sensor data.*
    - o Rationale: The sentence describes the technical aspects of an invention, specifically the use of two sensors to collect data, but it does not explicitly or implicitly mention any broader societal benefits. It only discusses the specific function of the two sensors and their relationship to each other, without indicating any public values.

- *on the other hand, human computer interfaces are not based on subtle human skills and are therefore cumbersome and unintuitive when compared to human spoken language and body language.*
    - o Rationale: The sentence discusses the limitations of human-computer interfaces in comparison to human communication, highlighting that interfaces are not based on subtle human skills and are therefore cumbersome and unintuitive. While this may imply a need for improvement in the user experience, it does not explicitly or implicitly relate to any public values or benefits. User experience enhancement do not imply public value.

- *an important task of autonomous devices is to avoid obstacles as the device moves around an environment, and to minimize risks of collision.*

- *in one or more embodiments, data provenance is collected, describing the execution of the various workflow tasks and the corresponding dataflow with features of input and output data sets of each task.*

**N2**. Do not code a sentence as a PVE if its benefits present only as accruing to the assignee organization (private value) without any explicit indication of broader societal benefit. Absent such explicit indication, do not code a sentence as a PVE by generalizing from private value to PVE based on inferences (by the coder) about potential connections. Avoid coder assumptions or inferences about potential (but not explicit) positive (or negative) societal, economic, or environmental spillovers.

*N2 examples.*

- *at times, data obtained from historical analytics may be applied to existing enterprise software systems, in order to ensure that operational activities of organizations are optimized to generate better results and minimize risk.*
    - o Rationale: This sentence is about a technology used to improve business performance. It is not possible to infer from this that the technology also contributes to public value. As far as the sentence conveys, value creation is limited to the enterprise user.

- *further, a good understanding of the labor market may assist a company deciding where to establish a new site because the company may choose a site with a readily available workforce.*
    - o Rationale: Something, perhaps a technology, helps the firm understanding the labor market. Such understanding assists the firm in its location decision. The only value reflected in the content of the sentence is that created for the firm. We cannot infer any public value from the sentence.

- *drivers with such reactions may exhibit unpredictable behavior, such as making attempts to counteract the semi-*

*autonomous functions, braking suddenly, or exhibiting other unpredictable and potentially dangerous behaviors.*

- *backend intelligence does not either collect data or use data available for analyzing the customer acceptance patterns and their choice of acceptable alternates.*

**N3**. Do not code a sentence as a PVE if it only refers to improving user experience. User experience enhancement does not imply a PVE. User experience enhancement focuses on individual users, is narrow in scope and individualistic, and is measured with usability metrics (e.g., task completion, user satisfaction). PVE focuses on society as a whole, is broader in scope and societal, and is measured with social impact metrics (e.g., environmental impact, social justice, democratic participation).

- *however, since types and strength of sensory effects desired by individual users may vary according to their user type which can be categorized by the user's gender, age, body, illness, and the like; current health state and mental state of the user; or a play environment (at dawn, an indoor or outdoor environment) of the sensory effects, standardized services which do not take into account the user preference cannot satisfy the user's requirements for experiencing diverse sensory effects.*
    - Rationale: The sentence discusses the customization of sensory effects based on user preferences, which is related to user experience enhancement. However, there is no explicit mention of broader societal benefits or public values, such as sustainability, social justice, or civil liberties. Additionally, the focus is on individual users and their preferences, rather than society as a whole. Therefore, this sentence does not meet the criteria for a Public Value Expression.

- *for example, if a user enters, via a user interface, a command for the autonomous floor cleaning system to "clean kitchen," without the user being able to confirm via a rendering of the environment that the autonomous floor cleaning system knows where the kitchen is, the autonomous floor cleaning system may clean the wrong part of the environment or otherwise misunderstand the command.*

- Rationale: While the sentence discusses a potential problem with the autonomous floor cleaning system's understanding of user commands, it primarily focuses on the individual user's experience and does not explicitly or implicitly reference broader societal benefits.